\documentclass[11pt]{article}

\usepackage{arxiv}

\usepackage[utf8]{inputenc} 
\usepackage[T1]{fontenc}    
\usepackage{hyperref}       
\usepackage{url}            
\usepackage{booktabs}       

\usepackage[export]{adjustbox}
\usepackage{pgfplots}
\pgfplotsset{width=12cm,compat=1.5}
\usepackage[mode=buildmissing]{standalone}

\usepackage{tikz}
\usepackage{amsmath} 
\usepackage{amssymb} 
\usepackage{amsfonts}

\title{Improving Vehicle Re-Identification using CNN Latent Spaces: Metrics Comparison and Track-to-track Extension}

\author{
  Geoffrey Roman-Jimenez \\
    Institut de Recherche en Informatique de Toulouse\\
    Université de Toulouse 3 \\
    Toulouse, France \\
    \& \\
    Solution Data Group, IA department \\
    Toulouse, France \\
    \texttt{groman-jimenez@solutiondatagroup.fr}\\
    \And
    Patrice Guyot   \\
    Institut de Recherche en Informatique de Toulouse\\
    Université de Toulouse 3 \\
    Toulouse, France \\
    \& \\
    École nationale supérieure des mines d'Alès \\
    Alès, France \\
    \And
    Thierry Malon \\
    Institut de Recherche en Informatique de Toulouse\\
    Université de Toulouse 3 \\
    Toulouse, France \\
    \And     
    Sylvie Chambon \\
    Institut de Recherche en Informatique de Toulouse\\
    Université de Toulouse 3 \\
    Toulouse, France \\
    \And
    Vincent Charvillat \\
    Institut de Recherche en Informatique de Toulouse\\
    Université de Toulouse 3 \\
    Toulouse, France \\
    \And
    Alain Crouzil \\
    Institut de Recherche en Informatique de Toulouse\\
    Université de Toulouse 3 \\
    Toulouse, France \\
    \And
    André Péninou \\
    Institut de Recherche en Informatique de Toulouse\\
    Université de Toulouse 3 \\
    Toulouse, France \\
    \And
    Julien Pinquier \\
    Institut de Recherche en Informatique de Toulouse\\
    Université de Toulouse 3 \\
    Toulouse, France \\
    \And
    Florence Sedes \\
    Institut de Recherche en Informatique de Toulouse\\
    Université de Toulouse 3 \\
    Toulouse, France \\
    \And
    Christine Sénac \\
    Institut de Recherche en Informatique de Toulouse\\
    Université de Toulouse 3 \\
    Toulouse, France \\}

\begin{document}
\maketitle

\begin{abstract}
This paper addresses the problem of vehicle re-identification using distance comparison of images in CNN latent spaces.

Firstly, we study the impact of the distance metrics, comparing performances obtained with different metrics: the minimal Euclidean distance ($MED$), the minimal cosine distance ($MCD$), and the residue of the sparse coding reconstruction ($RSCR$).
These metrics are applied using features extracted from five different CNN architectures, namely ResNet18, AlexNet, VGG16, InceptionV3 and DenseNet201. We use the specific vehicle re-identification dataset VeRi to fine-tune these CNNs and evaluate results. In overall, independently of the CNN used, $MCD$ outperforms $MED$, commonly used in the literature. These results are confirmed on other vehicle retrieval datasets.

Secondly, we extend the state-of-the-art image-to-track process (I2TP) to a track-to-track process (T2TP). The three distance metrics are extended to measure distance between tracks, enabling T2TP. We compared T2TP with I2TP using the same CNN models. Results show that T2TP outperforms I2TP for MCD and RSCR. T2TP combining DenseNet201 and $MCD$-based metrics exhibits the best performances, outperforming the state-of-the-art I2TP-based models.

Finally, experiments highlight two main results: i) the impact of metric choice in vehicle re-identification, and ii) T2TP improves the performances compared to I2TP, especially when coupled with $MCD$-based metrics.

\end{abstract}

\keywords{Track-to-track \and Vehicle re-identification \and Distance metrics \and Transfer Learning \and Deep latent representation \and Deep neural networks}

\section{Introduction}
\label{sec:intro}

With the recent growth of Closed-circuit Television (CCTV) systems in big cities, object re-identification in video surveillance, such as vehicle and pedestrian re-identification, is a very active research field. 
In the last few years, major progress has been observed in the vehicle re-identification field thanks to recent advances in machine- and deep-learning~\cite{khan2019survey}. 
These advances are very promising for intelligent video-surveillance processing, intelligent transportation and future smart city systems. 

Vehicle re-identification, in video surveillance, aims at identifying a query vehicle, filmed by one camera, among vehicles filmed by other cameras of a CCTV system. 
It relies on a comparison between a query vehicle and a database of known vehicles, to find the best matches. 
Commonly, the query is a single image and the vehicles of the database are represented by an image or a set of images called \emph{track}, extracted from video segments recorded by CCTV cameras.

In the literature~\cite{khan2019survey,feris2012large,zapletal2016vehicle,liu2016large,liu2016deep,liu2018provid, shen2017learning,liu2016vehicleid,liu2018ram,zhu2019vehicle, wu2019vehicle, de2019two, cui2017vehicle, bai2018group, peng2019learning}, vehicle re-identification is generally conducted as follows.
First, query and gallery vehicles are placed in a common space, by extracting features, representing the visual characteristics of the vehicle within one or several images, in order to share the same dimensions and be comparable to each other. Additionally, these features can be augmented using additional annotations (license plate, trend of the car, color of the car, etc.) and/or contextual metadata (camera location, time, road map, etc.).
Second, using a distance metric (or similarity) between these features, the gallery vehicles are ranked with respect to the query vehicle, from the first candidate to the last. 
Depending on the study, authors considered either an image-to-image process (I2IP) or an image-to-track process (I2TP) for the ranking. In I2IP, all images are ordered such that the ranking contains every image of each vehicle. In I2TP, the ranking only take the nearest image of each vehicle track as a reference. 

Previous studies have focused on the problem of feature extraction.
Feris \textit{et al.}~\cite{feris2012large} originally proposed an attribute-based method for vehicle re-identification using several semantic attributes (such as the category of vehicle and color).
Zapletal \textit{et al.}~\cite{zapletal2016vehicle} proposed to use color histograms and histograms of oriented gradients on transformed images (placing them in a common space) and a trained SVM classifier to perform vehicle re-identification.
Liu et al. \cite{liu2016large} were the first to evaluate and to analyze the use of Convolutional Neural Networks (CNNs) for vehicle re-identification, extracting the Latent Representation (LR) of the vehicles within the latent space of CNNs. They also provided a specific large-scale dataset for this purpose: the VeRi dataset. They evaluated the vehicle re-identification performance of LR extracted from several CNN architectures, and compared them to texture-based and color-based features. They showed that i) LRs of CNN architectures were particularly suitable for vehicle re-identification and ii) a linear combination of the three types of features was performing better.
Later, they showed that adding contextual information (license plate and spatiotemporal metadata) improves performance~\cite{liu2016deep, liu2018provid}.
Cui \textit{et al.}~\cite{cui2017vehicle} also proposed to fuse the LR of CNNs specialized in the detection/classification of vehicle details such as color, model, and pasted marks on the windshield. 
In the same vein, Shen \textit{et al.}~\cite{shen2017learning} incorporated complex spatiotemporal information to improve the re-identification results. 
They used a combination of a Siamese-CNN and a Long-Short-Term-Memory (LSTM) model to compute a similarity score, used for vehicle re-identification.
Instead of training a CNN to classify vehicles, Liu \textit{et al.}~\cite{liu2016vehicleid} suggested to directly learn a distance metric using a triplet loss function to fine-tune a pre-trained CNN. They also provided another large dataset containing a high number of vehicles, called vehicleID. 
Liu \textit{et al.}~\cite{liu2018ram} introduced a CNN architecture that jointly learns LRs of the global appearance and of local regions of the car. Attribute features (colors, model) are additionally used to jointly train their deep model. Finally, they concatenated global LR, local LR and attribute features. They concluded that the more information is combined, the higher the re-identification performance is. As an alternative of attribute combination for LR-based re-identification, De Oliviera \textit{et al.}~\cite{de2019two} used a two-stream siamese neural network in order to fuse information from patches of the vehicle shape's and patches of license plate. 
Using a multi-view approach, Huang \textit{et al.} \cite{huang2019multi} increased the re-identification performances by combining the information of consecutive frames of the same vehicle with the estimation of its orientation and metadata attributes. 
Questioning the \textit{transferability} of attribute-enriched models, Kabani \textit{et al.} \cite{kanaci2017vehicle} argued that the use of visual-only LR remains more flexible while achieving comparable results. 
Focusing on the development of more effective LR of vehicles, Zhu \textit{et al.}~\cite{zhu2019vehicle} fused quadruple directional deep features learned by using quadruple directional pooling layers, and were able to outperform most of the state-of-the-art methods without using extra vehicle information. Recently, using generative adversarial network, Wu \textit{et al.}~\cite{wu2019vehicle} proposed to generate unlabeled samples and a re-ranking strategy to boost the re-identification performances of off-the-shelf CNNs. Using also a re-ranking optimization, Peng \textit{et al.}~\cite{peng2019learning} increased the performances through a multi-region model that fuses features from global vehicle appearance and local regions of the vehicle images.

In these studies, the matching process uses the Euclidean distance, or a similarity score derived from it, to measure the distance between the query and a gallery vehicle image.
However, the use of Euclidean distance has often been criticized for being not well suited to high-dimensional spaces~\cite{domingos2012few}, such as those constructed by CNNs (often generating a dimension of features greater than 500). 
To our knowledge, the impact of the metric choice on the vehicle re-identification performances has not been addressed; this is the first issue addressed in this paper.

Furthermore, the systematic evaluation of distance metrics leads us to consider a more general framework than the commonly-used I2IP/I2TP which relies on image-to-image/image-to-track distance comparisons. 
Indeed, in the practice of vehicle re-identification, the query vehicle is selected directly on the video segment recorded from the camera of the CCTV system. This video segment provides a variety of valuable information that remains unused in I2IP/I2TP. For instance, in the case of a moving car, the video segment may offer different visual cues from the same vehicle (angle of view, zoom, brightness/contrast changes, etc.). This additional knowledge about the visual aspect of the query vehicle may improve the re-identification. Moreover, the use of a whole video segment may avoid the selection of only one specific query image without knowing the potential impact of such selection in the re-identification performances.  
The literature on vehicle detection and tracking is very rich, and numerous methods are today available to perform automatic vehicle detection and tracking in a given camera~\cite{swathy2017survey}. Therefore, assuming that the video segment selected by the user has to be processed by such algorithms, the query vehicle could be represented by a track, which would provide more information for the re-identification.
So far, the use of a query containing more than one image has not been fully addressed in vehicle re-identification. We address this issue by considering the track-to-track process (T2TP).

In this paper, we propose to i) evaluate the impact of the metric choice in re-identification and ii) extend the vehicle re-identification to T2TP and assess the performances in comparison with I2TP.
To this extent, the main experiments in this paper are made using the VeRi dataset. Indeed, unlike other large-scale dataset, VeRi contains image-based tracks of vehicles, allowing performances comparison between I2TP and T2TP, as well as comparison of performances with state-of-the-art methods. Note that since I2IP is not based on the same ranking support than I2TP and T2TP, I2IP is not considered in these experiments.
In addition, we further investigated the impact of the metric in other I2IP-based vehicle retrieval tasks using three other large-scale datasets of the literature, namely the VehicleID~\cite{liu2016vehicleid}, CompCars~\cite{yang2015large} and BoxCars116k~\cite{sochor2018boxcars}.

Let us underline that this paper focuses on visual information-only re-identification processes: no extra or contextual information is used in the studied processes. 
It is worth noting that the goal of this article is not to provide another re-identification system, but rather to evaluate the impact of the metric choice in the re-identification performance, and the potential benefits of T2TP on state-of-the-art methods.

This paper is organized as follows.
After introducing the mathematical notations in Section~\ref{sec:vereid}, we present the distance metrics that we compare in terms of re-identification performance, in Section~\ref{sec:metrics}.
Section~\ref{sec:t2t} presents the extension of the re-identification to T2TP. 
Then, Sections~\ref{sec:experiments} and \ref{sec:results} respectively present the experiments conducted to evaluate the re-identification performance and the results obtained.
Finally, in Sections~\ref{sec:discussion} and \ref{sec:conclusion} we discuss our results, give some perspectives, and conclude.

\section{Vehicle re-identification}

\label{sec:vereid}

In this section, we present the problem of vehicle re-identification. First, we introduce the mathematical notations that cover state-of-the-art I2TP, and T2TP (the second being a generalization of the first). 
Then, we present the two-step method for vehicle re-identification considered in our experiments, namely the LR extraction and the matching and ranking process. 

\subsection{Notations and problem statement}

Let consider $\mathcal{C}=\{C_1, C_2, ..., C_{n_c}\}$, the set of $n_c$ cameras of a CCTV system, and $\mathcal{V} = \{V_1, V_2, ..., V_{n_v}\}$, the set of $n_v$ vehicles captured by the cameras in $\mathcal{C}$. Each vehicle of $\mathcal{V}$ is uniquely identified.
We denote $\mathcal{T}=\{T_1, T_2, ..., T_{n_t}\}$ the set of $n_t$ tracks captured by the cameras of $\mathcal{C}$, and stored in a database. A track $T_k$, captured by one camera of $\mathcal{C}$, is associated with one of the vehicles of $\mathcal{V}$ denoted $V_k$.
Since a vehicle can be recorded by multiple cameras, two tracks $T_i$ and $T_{l}$ (with $l\neq i$) can be associated with the same vehicle, such that $V_i = V_{l}$. A track $T_i=\{I_{i,1}, I_{i,2}, ..., I_{i,N_{i}}\}$ is a set composed of $N_{i}$ images, all representing the same vehicle $V_i$. Each image $I_{i,j}$ of $T_i$ is cropped within the frame of its corresponding video segment from where it has been recorded. Note that, in this paper, we do not consider the time of the capture of each image, so that the order of images in a track is not taken into account. 
Given a query track $T_q=\{I_{q,1}, I_{q,2}, ..., I_{q,N_{q}}\}$, representing the vehicle $V_q \in \mathcal{V}$, $V_q$ being unknown, the aim of vehicle re-identification is to find a track $T_r \in \mathcal{T}$ in which the vehicle $V_q$ appears.
It is worth noting that, in case of I2TP, the query track $T_q$ is only composed of one image $I_q$.

Figure~\ref{lrextraction} shows a general overview of the vehicle re-identification process considered in this paper.

The first step consists of extracting features characterizing the vehicles in the track images. The feature extraction process is presented in Section \ref{subsec:lrextract}.
Using these features, the second step aims at ranking the different tracks of $\mathcal{T}$ based on their distance to the query.
The matching and ranking process is presented in Section \ref{subsec:lrranking}.

\subsection{Latent representation extraction}
\label{subsec:lrextract}

\begin{figure}[!h]
\centering 
\includegraphics[width=\linewidth]{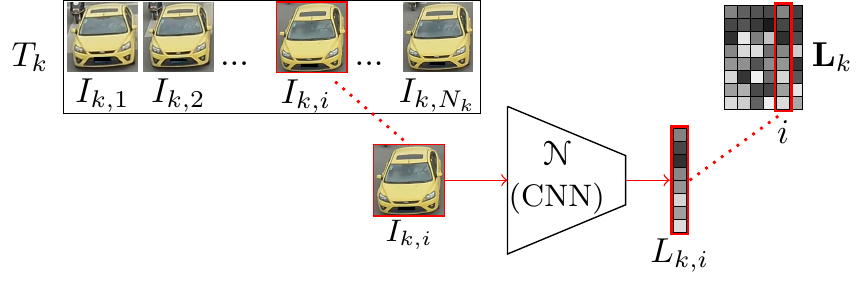}
\caption{Extraction of the latent representation $\mathbf{L}_k$ for a track $T_k$. Each image $I_{k,i} \in \mathbb{R}^{n \times m}$ of $T_k$ is transformed into a vector $L_{k,i} \in \mathbb{R}^f$ through the second-to-last layer of the CNN. The matrix $\mathbf{L}_k$ is then constructed as the concatenation of the $N_k$ vectors $L_{k,i}$.}
\label{fig:lrextraction_one}
\end{figure}

The aim of feature extraction is to represent all the images of each track of $\mathcal{T}$ in one common space, in order to make them comparable. 
We use as common space the latent spaces of CNNs and, as features, the latent representation (LR) of each image in these latent spaces. 
The main idea is to use one of the last layers of a CNN as a vector of features, in order to represent the input image in the latent spaces of the network. Formally, we consider a function $\mathcal{N}: \mathbb{R}^{n \times m} \rightarrow \mathbb{R}^f$ that transforms an image $I_{k} \in \mathbb{R}^{n \times m}$ to a vector of features $L_k \in \mathbb{R}^{f}$, $n\times m$ being the size of the image and $f$ being the dimension of the latent space. 
To represent the LR of a whole track, we concatenate each LR of its images to form a matrix.
Thus, we denote the matrix $\mathbf{L}_k=[L_{k, 1}, ..., L_{k, N_{k}}] \in \mathbb{R}^{f \times N_{k}}$, the LR of a track $T_k $, constructed as a concatenation of the LR of the $N_{k}$ images of the track. Similarly, the LR of a query track $T_q$ is denoted $\mathbf{L}_q \in \mathbb{R}^{f \times N_{q}}$. Let us notice that in case of I2TP, the LR of the query image $I_q$ is denoted $L_q$ $\in \mathbb{R}^{f}$.
Figure~\ref{fig:lrextraction_one} shows a graphical representation of the LR extraction for a track $T_k$. 

\subsection{Vehicle matching and ranking}
\label{subsec:lrranking}

Given a query track $T_q$, the aim of LR matching is to find the vehicle $V_{\tilde{r}}$, such that
\begin{equation}\label{eq:lrmatch}
    \tilde{r} = \underset{r}{\mathrm{argmin}}( d(\mathbf{L}_q, \mathbf{L}_r) ),
\end{equation}
with $r\in \{1, 2, ..., n_t\}$, and where $d$ is a distance function measuring how close the gallery track $T_r$ (represented by $\mathbf{L}_r$) is from the query track $T_q$ (represented by $\mathbf{L}_q$).

In order to evaluate the vehicle-re-identification, the matching process is conducted as a ranking on the gallery tracks, from nearest to farthest. This consists in ranking every track of $\mathcal{T}$ to construct an ordered set $\tilde{\mathcal{T}}_q = \{T_{q,1}, ..., T_{q,N_t}\}$, such that a track $T_{q,i}$ is the $i^{th}$ nearest track from the query according to the distance function $d(.)$, $T_{q, 1}$ being the first match (i.e. the nearest) and $T_{q, N_t}$, being the last (i.e. the farthest).

\begin{figure*}[!t]
\centering 
\includegraphics[width=\linewidth]{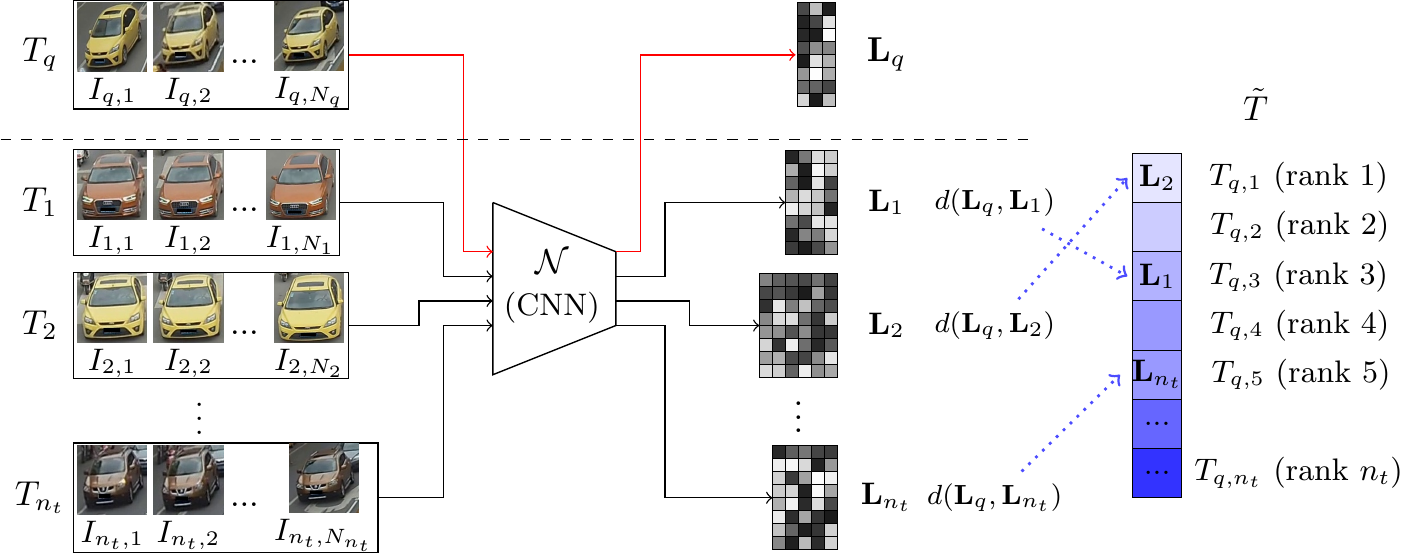}
\caption{Overview of the vehicle re-identification. Every vehicle image (query included) is represented by its LR within the latent space of the CNN. LR of all images of the same track are concatenated to build a matrix representing the LR of this track. Using a distance metric $d$, each track is ranked towards the query, from the closest to the farthest track, producing an ordered set $\tilde{T}$.}
\label{lrextraction}
\end{figure*}

\section{Image-to-track distance metrics}
\label{sec:metrics}

In this section we define the different distance metrics that we tested to compare their impact on vehicle re-identification. Referring to Figure~\ref{lrextraction}, we consider here a query track containing one image $T_q=I_q$ (represented by $L_q \in \mathbb{R}^{f}$) and a gallery track $T_r$ (represented by the vector $\mathbf{L}_r \in \mathbb{R}^{f \times N_r}$) taken from $\mathcal{T}$.

\subsection{Minimum Euclidean distance}

Euclidean distance has been widely used as a basic metric in many applications of content-based image retrieval~\cite{liu2007survey, wan2014deep}. In our context of vehicle re-identification, previous works only focused on the use of minimum Euclidean distance ($MED$) (or a variant)~\cite{feris2012large, zapletal2016vehicle, liu2016large, liu2016deep, liu2018provid,  shen2017learning, liu2016vehicleid, liu2018ram}. Therefore, we use $MED$ as a basis to evaluate the impact of other metrics (defined below) in the vehicle-re-identification.
We define the $MED$ function as:
\begin{equation}\label{eq:distmed}
    MED(L_q, \mathbf{L}_r) = \min_{i \in \{1, ..., N_r\} }(|| L_q - L_{r,i} ||_2),
\end{equation}
where $||.||_2$ is the $L_2$ norm measuring the Euclidean distance between the vector $L_q$ and a column of $\mathbf{L}_r$.

\subsection{Minimum cosine distance}
\label{subsec:cosim}

As a first alternative to $MED$, we propose to use the minimum cosine distance ($MCD$). Cosine distance is commonly used in data mining, machine learning~\cite{tan2006introduction}, and is often referred as being one of the most suitable distance metrics in information retrieval.
We compute the $MCD$ as follows:
\begin{equation}
     MCD(L_q, \mathbf{L}_r) = \min_{i \in \{1, ..., N_r\} }(1 - \frac{L_q^\top L_{r, i}}{|| L_q ||_2  || L_{r,i} ||_2} ), \\  
\end{equation}
where the term $\frac{L_q^\top L_{r, i}}{|| L_q ||_2 || L_{r,i} ||_2}$ corresponds to the cosine similarity between $L_q$ and $L_{r,i}$.
Note that, since we consider CNN architectures constructed with Rectified Linear Unit activation functions~\cite{nair2010rectified}, both elements of $L_q$ and $L_{r, i}$ are all positive.
Therefore, $MCD$ is bounded in $[0,1]$ (0 when $L_q=L_{r,i}$ and 1 when $L_q$ and $L_{r,i}$ are orthogonal).

\subsection{Residual of the sparse coding reconstruction}

Since Euclidean and cosine metrics are designed to measure distance between signals of the same dimension (here $\mathbb{R}^{f}$), these metrics are computed for each vector of $\mathbf{L}_r$ (corresponding to an image-to-image comparison). The minimum distance is then selected as the reference.
Therefore, among all images contained in tracks, at decision time, only one image is ever used to measure the distance between $L_q$ and $\mathbf{L}_r$.
To induce the use of more information, we propose to use the residual of the sparse coding reconstruction ($RSCR$). Sparse representation has been widely studied in many applications of computer vision, such as image classification, detection and image retrieval~\cite{zhang2015survey, wright2010sparse}.

We computed the $RSCR$ as follows: 
\begin{equation}\label{eq:i2trscr}
    RSCR(L_q, \mathbf{L}_r ) = {|| L_q -  \mathbf{L}_r\Gamma_{q,r} ||_2}^2,
\end{equation}
where $\Gamma_{q,r} \in \mathbb{R}^{N_r}$ is a code, combining linearly the column of the gallery $\mathbf{L}_r$, and optimized to reconstruct the query $L_q$ as follows:
\begin{equation}\label{eq:optimsc}
    \Gamma_{q,r} = \underset{\tilde{\Gamma}_{q,r}}{\mathrm{argmin}} ( {|| L_q -  \mathbf{L}_r \tilde{\Gamma}_{q,r} ||_2}^2  + \alpha || \tilde{\Gamma}_{q,r} ||_1).
\end{equation}
where $||.||_1$ is the $L_1$ norm maintaining the sparsity of the code, controlled by the coefficient $\alpha \in [0, 1]$.

\section{Extension to track-to-track re-identification}
\label{sec:t2t}

As an extension of I2TP, and referring to Figure~\ref{lrextraction}, T2TP aims at measuring the distance between a gallery track $T_r$ containing several images and the query track $T_q$. Here, LRs of $T_r$ and $T_q$ are respectively represented by $\mathbf{L}_r\in\mathbb{R}^{f\times N_r}$ and  $\mathbf{L}_q\in\mathbb{R}^{f \times N_q}$. 
Therefore, the main challenge with T2TP is to define metrics that are able to measure the distance between two tracks of different sizes.

\subsection{MED and MCD for T2TP} 

We extend the $MED$ and $MCD$ metrics to T2TP as follows.
First, considering a distance metric $d$ (e.g. $MED$ or $MCD$), we construct a set of distances $D_{q,r}= \{d(L_{q,j}, \mathbf{L}_r) \, | \, j \in {N_q}\}$ containing the $N_q$ computations of $d$ for each vector $j$ of $\mathbf{L}_q$ regarding $\mathbf{L}_r$. 
Then, we compute the overall distance between $T_q$ and $T_r$ by defining an aggregation function $g: \mathbb{R}^{n} \rightarrow \mathbb{R}$, in order to aggregate the elements of $D_{q,r}$, and obtain a scalar.

In our experiments, we used the following aggregation functions: \emph{minimum}, \emph{mean} and \emph{median}.
The \emph{minimum} function consists of selecting the best image-to-image match between the query and the gallery track, without taking into account the other images. Such function is therefore supposed to be more efficient when seeking for two tracks containing images with very similar points of view. The \emph{median} function also considers one image-to-image match, while promoting tracks containing at least half of its element similar to the query. On the contrary, the \emph{mean} function aggregates all elements of $D_{q,r}$, promoting tracks for which each image is similar to at least one image of the query, which can be sensitive to query with more variability.
With $d=MED$, we denote $minMED$, $meanMED$ and $medMED$ the T2TP metrics using respectively the aggregation function \emph{minimum}, \emph{mean} and \emph{median}. Similarly, with $d=MCD$, we denote the T2TP metrics, $minMCD$, $meanMCD$ and $medMCD$.
In addition, because some images of a track can be irrelevant for T2TP, we also consider the computation of truncated mean and median, using only the $N_q/2$ smallest distances within $D_{q,r}$. With $d=MED$, these metrics are denoted $mean50MED$ and $med50MED$. Similarly, with $d=MCD$, these metrics are denoted $mean50MCD$ and $med50MCD$.

\subsection{RSCR for T2TP}

Interestingly, since sparse coding is designed to reconstruct matrix, $RSCR$ can easily be extended to comply with track-based queries, by rewriting equations (\ref{eq:i2trscr}) to comply with $\mathbf{L}_q$:
\begin{equation}\label{eq:t2trscr}
    RSCR(\mathbf{L}_q, \mathbf{L}_r ) = || \mathbf{L}_q -  \mathbf{L}_r \mathbf{\Gamma}_{q,r} ||_F,
\end{equation}
where $||.||_F$ denotes the Frobenius norm, and where the sparse code $\mathbf{\Gamma}_{q,r} = [\Gamma_{q_1, r}, ..., \Gamma_{q_{N_q}, r}] \in \mathbb{R}^{N_r \times N_q}$ is computed by iteratively solving the equation (\ref{eq:optimsc}) for each column $\Gamma_{q_i,r} \in \mathbb{R}^{N_r}$ of $\mathbf{\Gamma}_{q,r}$, such that:
\begin{equation}\label{eq:t2toptimsc}
    {\Gamma}_{q_i,r} = \underset{\tilde{\Gamma}_{q_i,r}}{\mathrm{argmin}} ( {|| L_{q,i} -  \mathbf{L}_r \tilde{\Gamma}_{q_i,r} ||_2}^2  + \alpha || \tilde{\Gamma}_{q_i,r} ||_1).
\end{equation}

\subsection{Kernel distances}

As a natural extension of distance measurements between two sets of vectors (i.e. LR of tracks), we also propose to evaluate kernel distance metrics \cite{scholkopf2001kernel, phillips2011gentle}.
Kernel distance allows the measurement of the global distance between two tracks according to a given similarity kernel function $k$. The kernel distance $D_k$ between $\mathbf{L}_q$ and $\mathbf{L}_r$ is defined as : 
\begin{equation}\label{eq:kerneldist}
\begin{aligned}
    D_k^2(\mathbf{L}_q, \mathbf{L}_r) = & \sum_{i \in N_q}\sum_{j \in N_q} k(L_{q,i}, L_{q,j}) 
    \\
    & + \sum_{i \in N_r}\sum_{j \in N_r} k(L_{r,i}, L_{r,j}) \\
    & - 2\sum_{i \in N_q}\sum_{j \in N_r} k(L_{q,i}, L_{r,j}), 
\end{aligned}
\end{equation}
where $k(.)$ is a positive definite kernel function, measuring similarity between two vectors (here LR), such that $k(L_x, L_x) = 1$ and $k(L_x,L_y)$ decreases when the distance between $L_x$ and $L_y$ increases.

In our experiments, we tested two kernels, the radial basis function (RBF), defined as $k(L_x,L_y) = e^{\gamma {||L_x-L_y||_2}^2}$ (with $\gamma\in\mathbb{R}^+$, the \textit{spread} parameter of the function), and the cosine similarity (CoS), defined in Section \ref{subsec:cosim}. We respectively denoted these kernel distances $KRBF$ and $KCOS$.

\section{Experiments}
\label{sec:experiments}

We compared I2TP and T2TP performances and the impact of the metric by running experiments on the large-scale benchmark dataset VeRi~\cite{liu2016large}. 
We further evaluated the impact of the metric on other I2IT-based vehicle retrieval tasks VehicleID~\cite{liu2016vehicleid}, CompCars \cite{yang2015large} and BoxCars116k \cite{sochor2018boxcars}.

First of all, since the VeRi dataset is the only dataset containing several tracks for the same vehicle, we used it to evaluate the impact of the distance metric in I2TP and T2TP, as well as a performance comparison between them. We conducted our experiments on the VeRi dataset as follows. First, we used the training set of the VeRi dataset on five well-known CNN architectures to specialize them in the vehicle recognition task. We then used these fine-tuned CNNs to extract LR on every image of the testing set. Second, we evaluated I2TP and T2TP with respects to distance metrics defined in Section \ref{sec:metrics} and Section \ref{sec:t2t}.

Second of all, we extended the evaluation of the impact of the distance metric on other vehicle-based retrieval tasks using the datasets VehicleID, CompCars and BoxCars116k. These experiments were conducted using the DenseNet201 CNN architecture, which showed to be the best CNN architecture found during VeRI experiments.
With the VehicleID dataset we conducted two kinds of experiments, \textit{vehicle re-identification} and \textit{vehicle retrieval} tasks as originally proposed by the authors in \cite{liu2016vehicleid}. 
To evaluate the metric comparison to other LR-based retrieval tasks, we compared the impact of the metric on \textit{vehicle type recognition} task as in~\cite{sochor2018boxcars}, using the datasets BoxCars116k and CompCars. With the dataset BoxCars116k, since it contains the unique identifier of vehicles, we also conducted experiments of \textit{vehicle retrieval} as for VehicleID experiments.

\subsection{Experiments on the VeRi dataset: I2TP and T2TP comparison and impact of the distance metric }

\subsubsection{The VeRi dataset}

The VeRi dataset is composed of 49357 images of 776 vehicles recorded by 20 cameras in a real-world traffic surveillance system. Every vehicle of the dataset has been recorded by several of the 20 cameras of the system, constituting a totality of 6822 tracks of vehicles (each track is composed of a mean number of 6 images, varying from 3 to 14 images).
The VeRi dataset is divided into two sets, a training set, composed of 37778 images representing 576 vehicles (5145 tracks), and a testing set, composed of 11579 images representing 200 vehicles (1677 tracks). 
The training set is used to fine-tune the CNN for the task of \textit{vehicle recognition} as explained in section~\ref{subsec:finetune}. 
Evaluation of I2TP is performed through 1677 query images preselected in each track of the testing set.
Evaluation of T2TP is conducted using the 1677 tracks of the testing subset.
Since I2TP and T2TP both rely on the comparison of a query (that is either a unique image from a track or the whole track, taken from the testing set) to all other tracks of the testing set, their performances remain comparable.

\subsubsection{CNN architectures and LR extraction}
\label{subsec:cnnlrextract}

To extract LR, we used the second-to-last layer of popular CNN architectures, namely ResNet18~\cite{he2016deep}, VGG16~\cite{simonyan2014very}, AlexNet~\cite{krizhevsky2012imagenet}, InceptionV3~\cite{szegedy2016rethinking} and DenseNet201~\cite{huang2017densely} pre-trained on the dataset ImageNet~\cite{deng2009imagenet}.
These architectures, widely analyzed~\cite{zheng2018sift, alom2019state} and easily accessible \cite{paszke2017automatic}, have been chosen as a basis to evaluate the impact of the metrics and to compare I2TP and T2TP.

In order to comply with the inputs dimension of these CNNs, every image of the VeRi dataset was resized to $224\times224$. The different dimensions of the second to the last layer of ResNet18, VGG16, AlexNet, InceptionV3 and DenseNet201 are respectively 512, 4096, 4096, 2048 and 1920.

\subsubsection{Fine-tuning for vehicle classification}
\label{subsec:finetune}

To fine-tune the CNN models, we proceed as follows. 
We replaced the last layer of each CNN architecture by a fully-connected layer of 576 neurons, and trained each network to classify the 576 vehicles of the VeRi training set. The back-propagation was performed using the cross-correlation loss function. Weight optimization was performed using classical stochastic gradient descent (learning rate set to 0.001, momentum set to 0.9). The network was trained during 50 epochs.

\subsubsection{Evaluation protocol}

To evaluate the vehicle ranking, we use the Cumulative Matching Characteristic (CMC) curve which is widely used in object re-identification \cite{liu2016large, liu2016deep}. 
We reported the two measures rank1 and rank5 of the CMC curves, corresponding respectively to the precision at ranks 1 and 5.

Regarding the dataset VeRi, since there are several tracks that correspond to the query, we also computed the mean average precision (mAP) 
which is classically used in vehicle re-identification evaluation. 
mAP takes recall and precision into account to evaluate the overall vehicle re-identification.
Given a query $q$ and a resulting ranked set $\tilde{\mathcal{T}}_q$, the average precision (AP) is computed as
\begin{equation}
    AP(q) = \frac{1}{N_{gt}}  \sum_{k=1}^{N_t} \Big( \delta(T_{q, k}) \sum_{i=1}^{k}  \frac{\delta(T_{q, i})}{k} \Big) ,
\end{equation}
where $\delta(T_{q,i})$ is a function equal to 1 if the track $T_{q,i}$ represents the vehicle $V_q$, or 0 otherwise. $N_{gt}$ is the number of tracks representing the query vehicle $V_q$.

We computed mAP as the mean of all AP computed for every query:
\begin{equation}
    mAP = \frac{1}{N_{Q}} \sum_{q=1}^{N_Q} AP(q),
\end{equation}
with $N_Q$ being the number of queries performed with the dataset ($N_Q=1677$ with the VeRi dataset).

\subsection{Experiments on VehicleID, BoxCars116k and CompCars datasets: impact of the distance metric in other vehicle retrieval tasks}

For the experimentation on VehicleID, BoxCars116k and CompCars, we used the DenseNet201 architecture.
As in Section \ref{subsec:cnnlrextract}, we extracted LR from the second-to-last layer of the DenseNet201, constructing a LR of size 1920 for every image of the dataset.
For each dataset experiment, we fine-tuned the CNN for classification using their respective training set (described below), following the procedure described in Section \ref{subsec:finetune}.

Since these datasets do not contain several tracks for each vehicle, experiments are using I2IP for the ranking. Thus, in these experiments, $MED$ and $MCD$ respectively correspond to Euclidean distance and cosine distance. Furthermore, due to vector normalization applied for the resolution of sparse coding (Section \ref{subsec:impdetail}), and because squared difference between two normalized vectors is proportional to the cosine distance~\cite{choi2014toward}, $RSCR$ and $MCD$ will produce the same ranking. Therefore, $RSCR$ is not included in these experiments.

\subsubsection{Experiment on the VehicleID dataset}

The VehicleID dataset contains 221763 images of 26267 vehicles (each vehicle is represented by 8.42 images in average).
It is divided into two sets, a training set, composed of 113346 images representing 13164 vehicles, and a testing set, composed of 108221 images representing 13164 vehicles. To evaluate the effect of the scale of the dataset on retrieval performances, 3 subsets are extracted from the testing set: the \textit{Small} subset composed of 800 vehicles (6493 image), the \textit{Medium} composed of 1600 vehicles (13377 images) and the \textit{Large} subset composed of 2400 vehicles (19777 images). 
For both \textit{vehicle re-identification} and \textit{vehicle retrieval} experiments, the training set is used to fine-tune the CNN for the task of \textit{vehicle classification} (with 13164 classes).

For \textit{vehicle re-identification}, in each subset (\textit{Small}, \textit{Medium} and \textit{Large}), an image of each vehicle is randomly selected as a gallery image and the other images are used as query images, resulting in 5693, 11777, and 17377 query images. In this experiment, since only one gallery image correspond to the query image, only rank1 and rank5 are reported. 

Regarding \textit{vehicle retrieval}, for a given vehicle containing $N_t$ images, $max(6, N_t -1)$ are selected as gallery images and the rest as query images as in \cite{liu2016vehicleid}. For evaluation, rank1, rank5 and mAP measures are reported.

\subsubsection{Experiment on the BoxCars116k dataset}

The BoxCars116k dataset is composed of 116286 images of 27496 unique vehicles of 693 different vehicle models (brand, model, submodel, model year) collected from 137 different CCTV cameras with various angle viewpoints. BoxCars116k has been originally designed for fine-grained vehicle classification and vehicle type recognition~\cite{sochor2018boxcars}. For this purpose, authors constructed a subset, named "hard", containing 107 fine-grained vehicle classes (precise type of vehicle, including the model year) with uniquely identified vehicle divided into a training set of 11653 tracks (51961 images) and a testing set of 11125 tracks (39149 images). We used this subset for \textit{vehicle type recognition} and \textit{vehicle re-identifications} tasks. 

For \textit{vehicle type recognition}, we used the training set to fine-tuned the CNN considering the 107 classes of fine-grained vehicle models. Using the testing set, for all image of a given vehicle model (107 classes) we randomly selected an image as query and the rest as gallery images.

For \textit{vehicle re-identification} task, we used the 11653 unique vehicle identities as classes to fine-tuned the CNN. For testing, for a given track of vehicle, we randomly selected an image as query and the rest as gallery images. 

\subsubsection{Experiment on the CompCars dataset}

The CompCars dataset is composed of 214345 images of 1687 vehicles collected from the web and urban surveillance cameras. For our experiment, we used the Part-I and the "surveillance data" subsets of CompCars defined by the authors in \cite{yang2015large}. Part-I subset contains 30955 images of 431 vehicle models; the training set and testing set contain respectively 16016 and 14939 images of the same 431 vehicle models. The subset "surveillance data" is composed of 44481 images of 281 car models captured in the front view; the training set and testing set contains respectively 31709 and 13894 images of the same 281 car models. For both subsets, we used the training set to fine-tune the CNN for \textit{fine-grained vehicle classification} task considering the vehicle models as classes (431 classes for PART-I and 281 classes for the "surveillance data"). For testing, given a testing set and for all images of a given vehicle model, we randomly selected an image as query and the rest as gallery images.

\subsection{Implementations details}
\label{subsec:impdetail}

CNN architecture construction and training have been implemented using the Pytorch framework in Python~\cite{paszke2017automatic}.
Regarding the $RSCR$, we solved equations~(\ref{eq:optimsc}) and (\ref{eq:t2toptimsc}) by using the lasso-LARS algorithm (Lasso model with a regularization term $L_1$, fitted with Least Angle Regression)~\cite{Efron04leastangle}, with $\alpha=1$. 
We computed the kernel distance $KRBF$ with $\gamma=\frac{1}{f}$, $f$ being the LR dimension of the considered CNN. Distance metric computations were implemented using the package \textit{scikit-learn} in Python.
Source codes for LR extraction (Section \ref{subsec:lrextract}), distance metric computations (Sections \ref{sec:metrics} and \ref{sec:t2t}) and vehicle ranking (Section~\ref{subsec:lrranking}) are available at \url{https://github.com/GeoTrouvetout/Vehicle_ReID}.

\section{Results}

\label{sec:results}

\subsection{Results on the VeRI dataset}
\subsubsection{Image-to-track results}

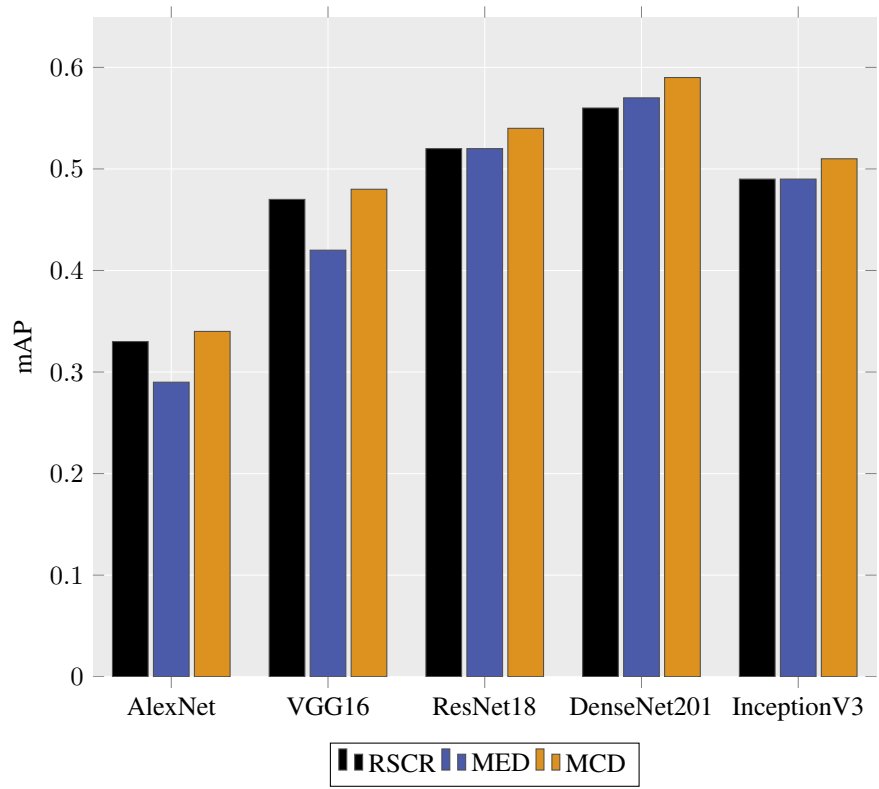
\begin{figure}
    \centering
    \definecolor{yellowgold}{RGB}{221, 146, 32}
    \definecolor{deepblue}{RGB}{75, 91, 169}
    \definecolor{greygrid}{RGB}{235, 235, 235} 
    \begin{tikzpicture}
    
    \begin{axis}[
    ybar,
    enlarge x limits=0.125,
    legend style={at={(0.5,-0.1)},
    	anchor=north,legend columns=-1},
    ymin=0,
    grid, 
    axis line style={draw=none},
    ylabel={mAP},
    bar width=3ex,
    symbolic x coords={AlexNet,VGG16,ResNet18, DenseNet201, InceptionV3},
    xtick=data,
    grid style={white},
    axis background/.style={fill=greygrid}
    ]
    \addplot[white!30!black, fill=black] coordinates {(AlexNet,0.33) (VGG16,0.47) (ResNet18,0.52) (InceptionV3,0.49) (DenseNet201,0.56) };
    \addplot[white!30!black, fill=deepblue] coordinates {(AlexNet,0.29) (VGG16,0.42) (ResNet18,0.52) (InceptionV3,0.49) (DenseNet201,0.57)};
    \addplot[white!30!black, fill=yellowgold] coordinates {(AlexNet,0.34) (VGG16,0.48) (ResNet18,0.54) (InceptionV3,0.51) (DenseNet201,0.59)};
    \legend{RSCR,MED,MCD}
    \end{axis}
    
    \end{tikzpicture}
    \caption{Image-to-track mAP results depending on the CNN architecture and the distance metrics used. The higher, the better.}
    \label{fig:i2tcurves}
\end{figure}

Table~\ref{tab:resi2t} reports the performances obtained with the metrics tested in I2TP ($MED$, $MCD$ and $RSCR$), depending on the CNN (AlexNet, VGG16, ResNet18, DenseNet201 and InceptionV3). 
Figure~\ref{fig:i2tcurves} depicts the mAP results obtained.

In terms of mAP, $MCD$ outperforms $MED$ for every CNN models (ranging from +2.02\% to +5.79\%). $RSCR$ outperforms $MED$ when associated with AlexNet (+3.74\%) and VGG16 (+4.82\%), but remains similar to ResNet18 (+0.87\%), InceptionV3 (-0.12\%) and DenseNet201 (-0.97\%). 
Overall, the best mAP result is obtained with DenseNet201 and $MCD$ (58.60\%). 

Regarding results of rank1 and rank5, $MCD$ outperforms $MED$ with AlexNet (rank1 +3.34\%, rank1 +3.28\%) and VGG16 (rank1: +3.94\%, rank5: +2.68\%), but performs similarly with ResNet18 (rank1: -0.06\%, rank5: +0.47\%), InceptionV3 (rank1: +1.13\%, rank5: +0.3\%) and DenseNet201 (rank1: -1.43\%, rank5: +0.3\%). $RSCR$ outperforms $MED$ when associated with AlexNet (rank1: +2.81\%, rank5: +2.8\%) and VGG16 (rank1: +3.58\%, rank5: +2.14\%), but performs slightly lower with other CNNs (rank1 ranging from -1.07\% to -2.51\%, rank5 ranging from -0.83\% to 0.18\%). 
Overall, the best rank1 is obtained with DenseNet201 and $MED$ (85.37\%), while the best rank5 is found with DenseNet201 and $MCD$ (95.41\%).

\begin{table*}[!t]
\caption{Image-to-track re-identification performance depending on the distance metrics and the CNN architecture used. Best performances are highlighted in bold.} 
\label{tab:resi2t}
\centering{
    \begin{tabular}{@{}ccccc@{}}

    \toprule CNN & Metric & mAP & rank1 & rank5\\ \midrule         
        & $MED$   & 29.08 & 63.98 & 81.1 \\
AlexNet & $MCD$   & 33.98 & 67.32 & 84.38 \\
        & $RSCR$ & 32.82 & 66.79 & 83.90 \\\midrule
        
        & $MED$  & 42.47 & 75.07 & 88.91 \\
VGG16   & $MCD$ & 48.26 & 79.01 & 91.59 \\
        & $RSCR$ & 47.29 & 78.65 & 91.05 \\ \midrule
        
        & $MED$  & 51.58 & 80.32 & 92.49 \\ 
ResNet18& $MCD$  & 53.66 & 80.26 & 92.96 \\  
        & $RSCR$  & 52.45 & 78.47 & 92.67 \\ \midrule

        & $MED$ & 56.58 & \textbf{85.57} & 95.11 \\
DenseNet201  & $MCD$ & \textbf{58.60} & 84.14 & \textbf{95.41} \\
        & $RSCR$ & 55.60 & 83.06 & 94.28 \\ \bottomrule
                
        & $MED$ & 48.91 & 77.28 & 91.53 \\
InceptionV3 & $MCD$ & 51.05 & 78.41& 91.83\\
        & $RSCR$ & 48.79 & 76.21 & 91.00\\ \midrule

\multicolumn{5}{l}{Note: Values are in percentages. The higher, the better.}

    \end{tabular}
    
}{}

\end{table*}

\subsubsection{Track-to-track results}

\begin{table*}[!t]
\centering{
\caption{Track-to-track re-identification performances (mAP, rank1 and rank5) depending on the metrics and the CNN architecture used. Best performances are highlighted in bold (RSCR, kernel distances, MED- and MCD-based metrics separately).}}
\label{tab:rest2t}
\centering{\scriptsize{
\begin{tabular}{@{}l l l l | lll | lll | lll | lll @{}}

\toprule
& \multicolumn{3}{c}{AlexNet} & \multicolumn{3}{c}{VGG16} & \multicolumn{3}{c}{ResNet18} & \multicolumn{3}{c}{DenseNet201} & \multicolumn{3}{c}{InceptionV3}   \\ \midrule
Metric    & mAP   & rank1 & rank5 & mAP   & rank1 & rank5 & mAP   & rank1 & rank5 & mAP   & rank1 & rank5 & mAP   & rank1 & rank5 \\ \midrule
$RSCR$  & 38.12 & 72.03 & 87.78 & 51.78 & 81.81 & 93.44 & \textbf{56.48} & 83.3  & 95.05 & 53.48 & \textbf{84.79} & \textbf{96} & 53.96 & 81.75 & 93.74 \\ \midrule
$KRBF$ & 12.99 & 42.75 & 53.25 & 36.02 & 75.73 & 86.46 & 40.33 & 79.55 & 88.61 &  53.14 & 84.62 & 92.67 & 48.59 & 80.44 & 90.52 \\
$KCOS$ & 31.03 & 68.28 & 84.91 & 46.58 & 79.49 & 90.7 & 53.76 & 82.71 & 92.55 & \textbf{54.45} & \textbf{84.91} & \textbf{93.14} & 51.44 & 80.92 & 91.41 \\ \midrule
$minMED$    & 29.6  & 63.21 & 80.56 & 42.91 & 75.13 & 89.03 & 55.43 & 83.84 & 94.69 & 60.7  &\textbf{88.97} & 96.72 & 52.33 & 81.45 & 94.04 \\
$meanMED$   & 25.89 & 59.63 & 78.59 & 40.28 & 74.12 & 87.84 & 54.58 & 83.42 & 94.81 & 58.48 & 87.66 & 96.24 & 50.84 & 80.98 & 92.61 \\
$medMED$    & 25.58 & 60.05 & 78.95 & 39.65 & 73.29 & 87.84 & 54.3  & 83.24 & 94.39 & 58.3  & 87.95 & 96.18 & 50.33 & 80.92 & 92.55 \\
$mean50MED$ & 28.33 & 62.85 & 80.8  & 42.5  & 75.43 & 88.91 & 56.13 & 84.73 & 95.23 & \textbf{60.37} & 88.49 & 96.66 & 52.53 & 82.41 & 93.62 \\
$med50MED$  & 27.89 & 62.43 & 80.32 & 42    & 74.78 & 88.55 & 55.63 & 84.38 & 94.93 & 60.07 & 88.43 & \textbf{96.9}  & 52.04 & 82.29 & 93.8  \\ \midrule
$minMCD$    & 38.4  & 71.79 & 87.95 & 52.83 & 82.65 & 94.57 & 58.08 & 84.2  & 95.53 & 62.31 & 87.06 & 96.78 & 55.48 & 83.06 & 94.69 \\
$meanMCD$   & 36.13 & 70.24 & 87.66 & 52.1  & 81.45 & 93.8  & 58.24 & 83.72 & 95.11 & 62.53 & 86.64 & 96.6  & 54.56 & 81.75 & 93.56 \\
$medMCD$    & 35.86 & 70.18 & 87.95 & 51.63 & 81.04 & 94.04 & 57.89 & 83.3  & 95.35 & 62.08 & 86.82 & 96.72 & 54.37 & 82.41 & 93.98 \\
$mean50MCD$ & 37.93 & 72.09 & 87.78 & 53.26 & 82.89 & 94.28 & 59.17 & 84.38 & 95.95 & \textbf{63.2}  & \textbf{87.36} & 97.02 & 55.94 & 83.84 & 94.69 \\
$med50MCD$  & 37.8  & 71.91 & 88.01 & 52.91 & 82.47 & 94.22 & 58.92 & 84.2  & 95.59 & 62.82 & 87.06 & \textbf{97.08} & 55.71 & 83.3  & 95.95 \\ \bottomrule

\multicolumn{16}{l}{Note: Values are in percentages. The higher, the better.}
\end{tabular}

}}

\end{table*}


\begin{figure*}[!t]
    \centering
    \includegraphics[width=\linewidth]{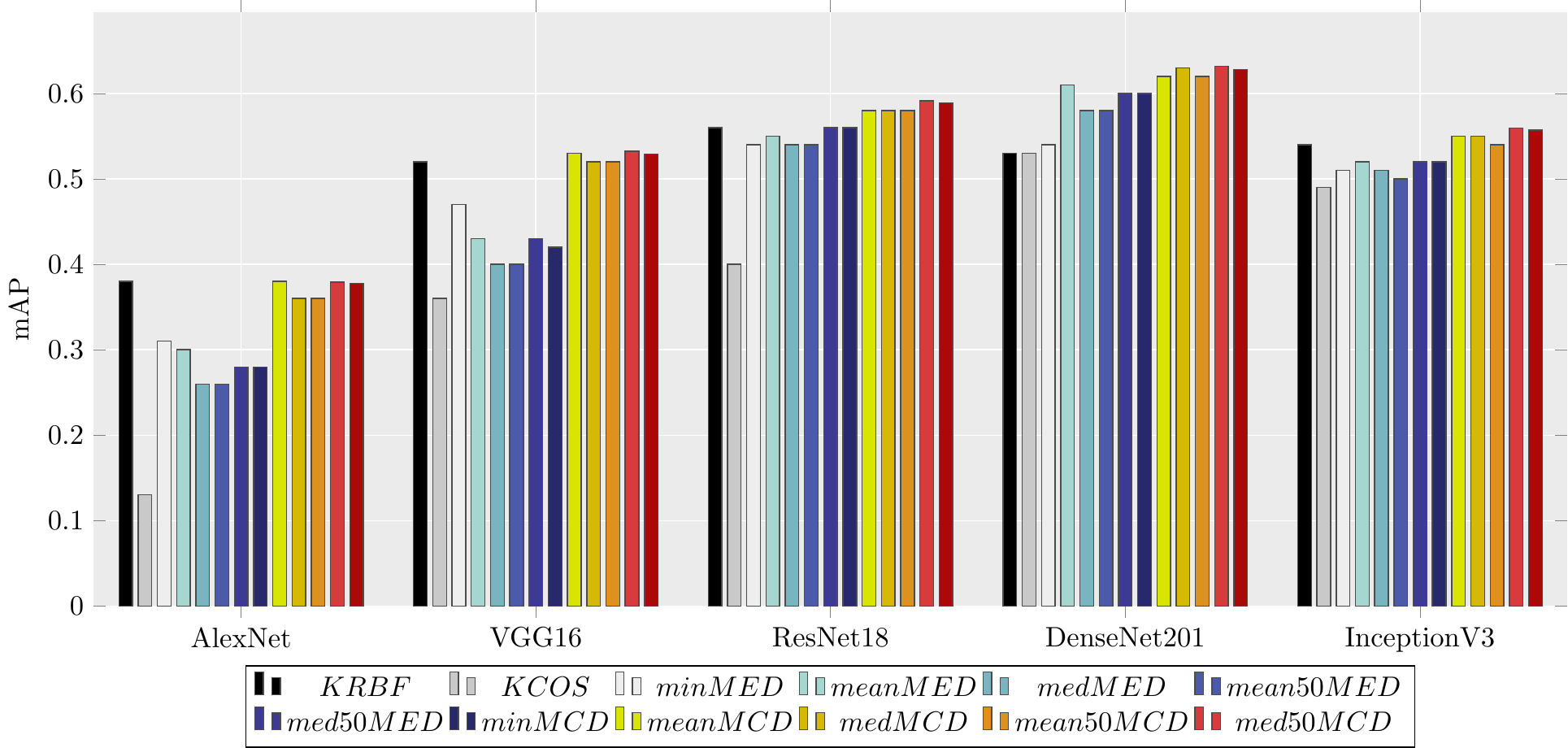}
    \caption{Track-to-track mAP results depending on the CNN architecture and the distance metric used. Black bars correspond to RSCR metric. Grey and white bars correspond to kernel distances, respectively KRBF and KCOS. Blue-colored bars represent the MED-based metrics. Warm colors (yellow to red) bars represent the MCD-based metrics. The higher, the better.}
    \label{fig:t2tcurves}
\end{figure*}

Table~\ref{tab:rest2t} reports the T2TP performances obtained with the different metrics tested ($RSCR$, $KRBF$, $KCOS$, $MED$- and $MCD$-based metrics), depending on the CNN. Figure~\ref{fig:t2tcurves} depicts the mAP results obtained. 

For each CNN taken individually, T2TP outperforms I2TP independently of the metric (with the exception of $KRBF$ and $KCOS$, not computed with I2TP). The gain of mAP is respectively +0.34\%$\pm$2.63 for the $MED$-based metrics, +4.07\%$\pm$0.85 with $MCD$-based metrics, and +3.37\%$\pm$3.11 for $RSCR$.

Comparing aggregation functions pairwise, $MCD$-based metrics outperform $MED$-based metrics independently of the CNN (mAP: +6.14$\pm$3.65\%). Both for $MED$- and $MCD$-based metrics, the aggregation function $mean50$ outperforms others.
Kernel distances ($KRBF$ and $KCOS$) performed poorly in comparison with $MED$- and $MCD$-based metrics.
With the exception of results obtained with DenseNet201, $RSCR$ outperformed $KRBF$ (mAP: +12.55\%$\pm$9.77\%) and $KCOS$ (mAP: +3.312\%$\pm$3.05\%). 
Overall, the different combinations of DenseNet and $MCD$-based metrics provide the best overall performance (mAP: [62.08\% -- 63.2\%], rank1: [86.64\% -- 87.36\%] and rank5: [96.6\% -- 97.08\%]). Best performance are found with DenseNet and $mean50MCD$ (mAP: 63.2\%, rank1: 87.36\%). 

\begin{figure*}[!t]
    \centering
    \includegraphics[width=\linewidth]{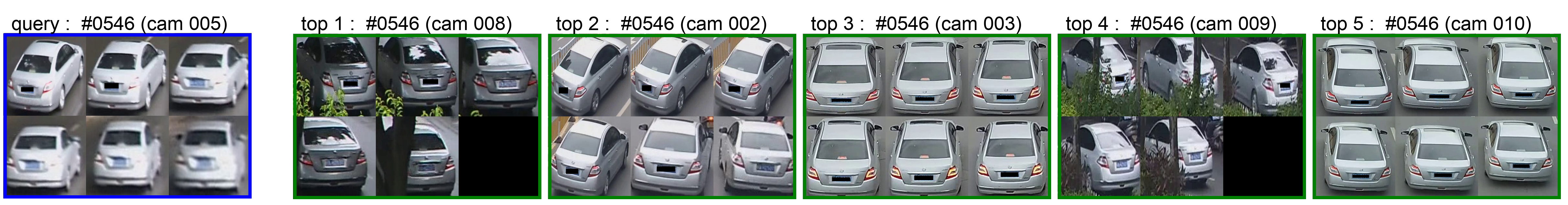}
    \includegraphics[width=\linewidth]{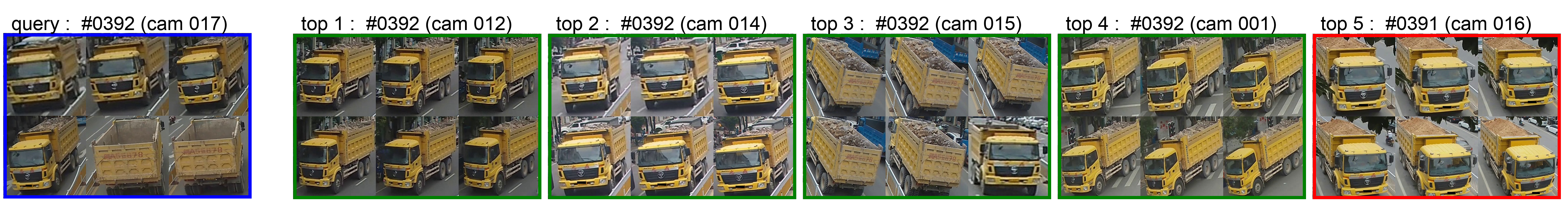}
    \includegraphics[width=\linewidth]{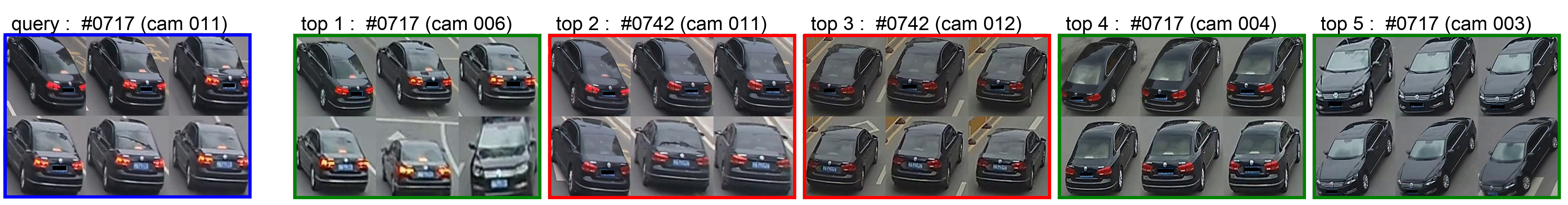}

    \caption{Qualitative examples of queries and ranking obtained with DenseNet in T2TP with $mean50MCD$. Each row indicates the query track (blue frame) and its corresponding top-5 ranking. Red frame indicates incorrect retrievals and green indicates correct retrievals. A maximum of 6 images per track are displayed.}    
    \label{fig:qualiex}
\end{figure*}

Figure~\ref{fig:qualiex} shows some visual results obtained with DenseNet and $mean50MCD$. In the first example (white car), we can observe that the model was able to correctly retrieve tracks containing images of the vehicle behind other elements (tree and bush) and with different angles of view.
The second and third examples (yellow truck carrying rocks and the black car) shows that the model was able to retrieve the correct vehicles, but was not able to distinguish between similar vehicles (a yellow truck carrying sand or another black car).
Other examples  of visual results are available at \url{https://cloud.irit.fr/index.php/s/cBWsTDBHfcWnJ9y}.

\subsubsection{Results on other dataset}

\subsubsection{Results on the VehicleID dataset}

Table~\ref{tab:resvehicleid} reports the performances obtained on the VehicleID dataset regarding the \textit{vehicle re-identification} and \textit{vehicle retrieval} tasks. 

Regarding the \textit{vehicle re-identification} task, for all the three subsets (small, medium, large), $MCD$ systematically outperforms $MED$ for rank1 ([+0.88\% -- +1.71\%]) and rank5 ([+0.75\% -- +1.35\%]).

Similarly, regarding the \textit{vehicle retrieval} task, $MCD$ outperforms $MED$ in terms of mAP ([+1.43\% -- +2.18\%]), rank1 ([-0.66\% -- +0.99\%]) and rank5 ([+0.31\% -- + 0.69\%]). 

\begin{table*}
\centering{
\caption{Distance metric comparison on the \textbf{VehicleID dataset}}
\label{tab:resvehicleid}

\begin{tabular}{@{}lccc|ccc@{}}
\toprule
\multicolumn{2}{c}{} & \multicolumn{2}{c}{Vehicle re-identification} & \multicolumn{3}{c}{Vehicle retrieval} \\ \midrule
Subset & Metric & rank1 & rank5 & mAP & rank1 & rank5\\ \midrule
small  & $MED$  & 62.83 & 71.37 & 66.64 & 95.73 & 96.95\\
       & $MCD$  & 64.54 & 72.12 & 68.07 & 95.12 & 97.26 \\ \midrule    
medium & $MED$  & 61.19 & 68.71 & 61.27 & 92.24 & 94.67 \\
       & $MCD$  & 62.07 & 69.65 & 63.05 & 93.00 & 94.98 \\ \midrule
large  & $MED$  & 58.80 & 67.13 & 60.19 & 90.80 & 94.46 \\
       & $MCD$  & 60.24 & 68.48 & 62.37 & 91.79 & 95.15 \\
       \bottomrule

\multicolumn{7}{l}{Note: Values are in percentages. The higher, the better.}
\end{tabular}
}{}   
\end{table*}

\subsubsection{Results on the BoxCars116k dataset}

Table~\ref{tab:resvehicleid} reports the performances obtained on the BoxCars116k dataset regarding the \textit{vehicle re-identification} and \textit{vehicle type recognition} tasks. 

For the \textit{vehicle retrieval}, $MCD$ systematically outperforms $MED$, with mAP (+1.95\%), rank1 (+1.78\%) and rank5 (+1.89\%).

For \textit{vehicle type recognition} task, $MED$ and $MCD$ reach high performances ([96.26\% -- 99.07\%]), and $MCD$ performed slightly better than $MED$ (mAP +0.96\%, rank1 +0.94\% and rank5 +0.94\%).

\begin{table*}
\centering{
\caption{Distance metric comparison on the \textbf{BoxCars116k dataset} regarding the \textit{vehicle re-identification} and \textit{vehicle type recognition} tasks}
\label{tab:resboxcars116k}

\begin{tabular}{@{}lcccc@{}}
\toprule
Task                 & Metric & mAP & rank1 & rank5 \\ \midrule
Vehicle              & $MED$ & 73.09 & 66.30 & 81.16 \\
re-identification    & $MCD$ & 75.04 & 68.08 & 83.05 \\ \midrule
Vehicle type & $MED$ & 97.08 & 96.26 & 98.13 \\
recognition          & $MCD$ & 98.05 & 97.20 & 99.07 \\ \bottomrule   
\multicolumn{5}{l}{Note: Values are in percentages. The higher, the better.} \\

\end{tabular}
}{}

\end{table*}

\subsubsection{Results on the CompCars dataset}

Table~\ref{tab:resvehicleid} shows the performances on the \textit{vehicle type recognition} task obtained with the two subsets "PART-I" and "surveillance data" of CompCars. 

Regarding the "PART-I" subset, $MCD$ outperformed $MED$ with +2.79\% for rank1 and +0.93\% for rank5. Reaching high performances with the "surveillance data" subset ([97.84\% -- 98.92\%]), $MCD$ performed slightly better than $MED$ (+0.35\% for rank1 and equally for rank5). 

\begin{table*}
\caption{Distance metric impact on \textbf{CompCars dataset} regarding the \textit{vehicle type recognition} task}
\label{tab:rescompcars}
\centering{
\begin{tabular}{@{}lccc@{}}
\toprule
Subset             & Metric & rank1 & rank5 \\ \midrule
PART-I             & $MED$  & 77.49 & 88.86 \\
(web-source data)  & $MCD$  & 80.28 & 89.79 \\ \midrule
surveillance data  & $MED$  & 97.84 & 98.92 \\
                   & $MCD$  & 98.20 & 98.92 \\ \bottomrule
\multicolumn{4}{l}{Note: Values are in percentages. The higher, the better.} \\
\end{tabular}
}{}   
\end{table*}

\section{Discussion and perspectives}
\label{sec:discussion}

From a general point of view, we can observe high variability of performance between CNNs. As expected, such results confirm the impact of the CNN architectures on the re-identification performance. This demonstrates the relevance of previous works focusing on the definition of specific CNN architectures and on the learning of efficient LR.

Besides, considering a given CNN architecture to produce LR, our results also show high variability of performance depending on the distance metric, showing that the choice of the metric for the matching process has a major impact on re-identification performance.

\subsection{Impact of the metric in image-to-track re-identification performances}

\subsubsection{Limitations of MED}

Globally, experiments on the VeRi dataset show a clear gain of performance from $MED$ to $MCD$ (mAP gain ranging from +2.02\% to +5.79\%). 
More precisely, we can observe big difference of performance between $MED$ and $MCD$/$RSCR$, especially when associated with AlexNet and VGG16.
This could be related to the higher dimension of the LR produced by these CNNs ($\mathbb{R}^{4069}$), potentially more affected by the \textit{curse of dimensionality}~\cite{verleysen2005curse}, compared to other CNNs ($\mathbb{R}^{512}$,  $\mathbb{R}^{1920}$ and $\mathbb{R}^{2048}$).
Therefore, besides the obvious differences of performance between CNN architectures, we argue that such dim\-en\-sio\-na\-li\-ty-per\-for\-mance relationship could have limited $MED$-based results in the literature. For instance, with their RAM architecture, Liu \textit{et al.}~\cite{liu2018ram} concatenated vectors of features into a single vector of dimension > 6000. Thus, we think that the use of $MED$ metric during their matching process may have reduced the performance of their system, which could be improved with a more appropriate metric (e.g. $MCD$).

\subsubsection{Performance of MCD}

Cosine measure has been shown to be a powerful metric when dealing with high dimensional features~\cite{ertoz2003finding}, in various applications~\cite{nguyen2010cosine, li2013distance}.
In our I2TP-based VeRi experiments, $MCD$ metric clearly outperforms $MED$ in terms of mAP, and remains similar regarding the metrics rank1 and rank5. This can be interpreted as the fact that $MCD$ provides overall better ranking of vehicles, improving the retrieval of other correct track of vehicles that are not in the first ranks, without impacting the retrieval of top-rank vehicle tracks. In addition, $MCD$ demonstrates adaptive capabilities to various dimensions of features (from  $\mathbb{R}^{512}$ to $\mathbb{R}^{4096}$).
Overall, the performances gain obtained with $MCD$ suggests that cosine-based metric can be considered as an interesting, and easy to implement, alternative to Euclidean-based metric (such as $MED$).

\subsubsection{Impact of the metric on other LR-based vehicle retrieval tasks} 

Experiments on VehicleID, BoxCars116k and CompCars also showed that $MCD$ systematically outperformed $MED$ on I2IP-based \textit{vehicle retrieval}, \textit{vehicle re-identification} and \textit{vehicle type recognition} tasks. 
In these experiments, rank1 is slightly more improved (gain ranging from +0.88\% to +2.79\%) than rank5 (gain ranging from +0.31\% to +1.35\%) using $MCD$ instead of $MED$, suggesting that $MCD$ is able to rank in the first positions more similar images than $MED$.
Overall, the systematic gain across each I2IP experiment suggests that the improvement of performances using $MCD$ over $MED$ could be generalized to other LR-based retrieval tasks.

\subsection{Performance improvement with T2TP}
From a general point of view, T2TP outperforms I2TP independently of the metric (with the exception of KRBF and KCOS, not computed with I2TP). The gain of mAP is respectively +0.34\%$\pm$2.63 for the $MED$-based metrics, +4.07\%$\pm$0.85 for the $MCD$-based metrics, and +3.37\%$\pm$3.11 for the $RSCR$. These results clearly illustrate the interest of using track-based query to help the re-identification process.
Obviously, such gain of performance had to be expected since a track-based query (T2TP) contains more visual information than an image-based query (I2TP).
Nevertheless, we can observe that the gain of performance is higher with $MCD$-based and $RSCR$ metrics than $MED$-based metrics (with the exception of DenseNet201 for $RSCR$). In addition, T2TP-specific metrics ($KRBF$ and $KCOS$) performed poorly compared to others, indicating that global track-to-track distance measurements, taking into account all the images of both tracks, seem to be less effective than more ``selective'' ones.
Thus, results outline that a significant improvement of performances with T2TP can only be obtained when combined with a relevant and adapted metric.


\subsubsection{Aggregation function}

As mentioned above, results show the extension of I2TP metrics to T2TP ($MED$- and $MCD$-based metrics) seem more effective than T2TP-specific metrics ($KRBF$ and $KCOS$).
However, the generalization of $MED$ and $MCD$ to T2TP is not straightforward, and induces, in the absence of \textit{a priori} knowledge on the vehicle tracks, an arbitrary choice of aggregation function. 
In our experiments, the aggregation functions $min$ and $mean50$ show the best overall performances. 
As $MED$ and $MCD$ in I2TP, the $min$ function consists in selecting the best image-to-image distance between all pairs of images, focusing the re-identification on the best possible match between the query and a gallery vehicle. Therefore, the performance obtained with this metric depends on the existence of similar images between tracks of the same vehicle. 
Alternatively, the aggregation function $mean50$ has the advantage of aggregating the distances between query and gallery track images, while truncating irrelevant images contained in the query track. Such aggregation function is thus supposed to be less dependent on the existence of similar images between tracks of the same vehicle. Nevertheless, since the VeRi dataset mainly contains tracks with similar images, such effects are hard to evaluate.

Further experiments including more diversity in tracks of vehicles are thus needed. For instance, the PKU-VD~\cite{yan2017exploiting} and ToCaDa~\cite{malon2018toulouse} datasets provide tracks of vehicles containing different points of view (e.g. a track containing images of the vehicle in front and side-view). Although these datasets are not meant to assess re-identification performances as VeRi, they could be used to evaluate the effect of using more diverse images over tracks (more viewpoints of the vehicles, lack of similar images, etc.), and hence evaluate the benefit of T2TP.

\subsubsection{Advantages of RSCR}
Despite the relatively poor results obtained with $RSCR$ (compared to outperforming $MCD$-based results), we think that the use of sparse coding reconstruction remains an interesting method to explore in the context of LR-based re-identification.
First, $RSCR$ has the advantage of being directly usable for both I2TP and T2TP, without having to define any arbitrary aggregation function (like $MED$- and $MCD$-based metrics), or to perform a global comparison between tracks (like kernel distances). 
Second, unlike other distance metrics, $RSCR$ is based on linear combinations (the sparse coding reconstruction) of LR, which are expected to induce complex semantic operations between the visual cues present in the images. Mikolov \textit{et al.}~\cite{mikolov2013efficient} in the domain of word representation and Radford \textit{et al.}~\cite{radford2015unsupervised} in synthetic image generation showed that simple arithmetic operations between objects in latent spaces of DNN can correspond to complex transformations between semantic concepts. 
In our context of vehicle re-identification, linear combination performed with $RSCR$ can be viewed as a combination between the various existing points of view of a given vehicle, which could potentially produce LRs corresponding to unseen points of view of the vehicle. Hence, in contrast to other metrics, $RSCR$ could be more robust to the absence of similar images between tracks. In addition, the sparse constraint narrows this linear combination to the most useful LRs, avoiding the use of irrelevant images (e.g. images of vehicle in back-view to retrieve a vehicle seen in a front-view, noisy images, etc.) and/or redundant information (e.g. stationary vehicle), in the reconstruction. 

Future work will focus on evaluating the advantages of using $RSCR$, and more generally of metrics based on linear combination of LRs, in the context of vehicle re-identification.

\subsection{Comparison with the state-of-the-art methods}

\begin{table*}[t]
\centering{
\caption {Comparison with the state-of-the-art methods on \textbf{VeRi dataset}.}
\label{tab:comparison}
    \begin{tabular}{@{}c c c c@{}}
    \toprule
    \textbf{Method} & mAP & rank1 & rank5\\ \midrule        
    BOW-SIFT~\cite{liu2018provid} & 1.51  & 1.91  & 4.53  \\
    LOMO~\cite{liu2018provid} & 9.41  & 25.33 & 46.48 \\  
    BOW-CN~\cite{liu2018provid} & 12.20 & 33.91 & 53.69 \\
    VGG~\cite{liu2018provid} & 12.76 & 44.10 & 62.63 \\    GoogleLeNet~\cite{liu2018provid} & 17.89 & 52.32 & 72.17 \\
    FACT~\cite{liu2018provid} & 18.75 &  52.21 & 72.88 \\
    nuFACT~\cite{liu2018provid} & 48.47 & 76.76& 91.42 \\
    RAM (baseline: only LR)~\cite{liu2018ram} & 55.0 & 84.8  & 93.1\\
    RAM~\cite{liu2018ram} & 61.5 & \textbf{88.60} & 94.00  \\
    QD\_DLF~\cite{zhu2019vehicle} & 61.83 & 88.50 & 94.46 \\ \midrule
    I2TP+Densenet201+$MCD$ & 58.60 & 84.14 & 95.41 \\
    T2TP+Densenet201+$mean50MCD$ & \textbf{63.2}  & 87.36 & \textbf{97.02} \\ \midrule
    GS-TRE \cite{bai2018group}* & 59.47 & \textbf{96.24} & \textbf{98.97} \\ 
    SSL\cite{wu2019vehicle}* & 61.07 &	88.57 & 93.56  \\
    SSL+re-ranking \cite{wu2019vehicle}* &\textbf{69.90} & 89.69 &95.41 \\
    MRM \cite{peng2019learning}* &	68.55 & 91.77 & 95.82 \\
    \bottomrule
    \multicolumn{4}{l}{Note: Values are in percentages. The higher, the better.} \\
    \multicolumn{4}{l}{*method using image-to-image process for the ranking} \\
    \end{tabular}
}{}
\end{table*}

We compared our best results (in I2TP and T2TP) with several recent methods including all the methods reported by Liu \textit{et al.} in~\cite{liu2018provid} (namely, BOW-SHIFT, LOMO BOW-CN, VGG, GoogLeNet, FACT and nuFACT), RAM~\cite{liu2018ram} (the baseline LR-only version of RAM is also reported), QD\_DLF~\cite{zhu2019vehicle}, GS-TRE~\cite{bai2018group}, SSL~\cite{wu2019vehicle} (with and without re-ranking), and MRM~\cite{peng2019learning}.
Performance comparison is summarized in Table~\ref{tab:comparison}. 

First, using only visual information (LR), the method combining DenseNet201 and $MCD$ (in I2TP) outperforms FACT and nuFACT~\cite{liu2018provid}, which use a combination of the visual aspect and contextual information. The method DenseNet201+$MCD$ also outperforms the state-of-the-art RAM ``baseline''~\cite{liu2018ram}, which only uses the global visual aspect of vehicles (like in our approach). These first results highlight the importance of the metric in the re-identification process, indicating that the use of $MCD$ is a more relevant metric than $MED$ in LR-based vehicle re-identification.

Second, the method combining DenseNet201 and $mean\-50\-MCD$ in T2TP outperformed the state-of-the-art RAM, QD\_DLF and GS-TRE methods~\cite{liu2018ram,zhu2019vehicle,bai2018group} in terms of mAP (respectively +1.35\%, +1.7\% and +3.73\%), but not the SSL+re-ranking method proposed by Wu \textit{et al.}~\cite{wu2019vehicle} and MRM proposed by Peng \textit{et al.}~\cite{peng2019learning}. 

However, these results should be balanced with the fact that authors of \cite{bai2018group}, \cite{wu2019vehicle} and \cite{peng2019learning} used I2IP for the ranking process instead of I2TP as used in~\cite{liu2018provid, liu2018ram, zhu2019vehicle}. 
In I2IP, each image of each vehicle is ordered individually, the ranking of I2IP and I2TP/T2TP are not based on the same support (images for I2IP, tracks of vehicles for I2TP/T2TP). Therefore, except for rank1 which only consider the first position of the ranking, performances between I2IP and I2TP/T2TP are difficult to compare. 

Considering the performance improvement obtained with only global visual information of vehicle images (no local features, no metadata/contextual information) and the very simplistic learning procedure that we used in our experiments (fine-tuning of standard CNN architectures), we argue that a relevant metric ($MCD$) combined with the use of more visual cues of the query vehicle (T2TP), could easily improve the performances of state-of-the-art methods which are specifically designed for vehicle re-identification.

\subsection{Limitation of LR visual-only based re-identification}

As stated and studied in~\cite{liu2016deep,liu2018provid, shen2017learning}, qualitative examples presented in Figure~\ref{fig:qualiex} confirm that visual-only based methods remain limited in their capacity to distinguish visually-similar vehicles. As an example, the model was not able to discriminate between two similar yellow trucks carrying respectively rocks and sand. This is possibly due to the use of global visual-only features, limiting the detection of details. To overcome such limitation, the use of region-based features, as in \cite{liu2018ram} and \cite{peng2019learning}, could allow the detection of small details differing between two similar vehicles, and increase the re-identification performances. 
In addition, visual-only based methods seem to hardly discriminate two similar cars with the same color and model (see the black car example of Figure~\ref{fig:qualiex}). In such case, the use of contextual metadata, such as spatiotemporal information and/or licence plate, as in \cite{shen2017learning} and \cite{liu2018provid}, is required to reach better discrimination between similar vehicles.

Finally, in this paper, we focused on the transfer learning approach which consists of reusing pre-trained CNN latent spaces to extract features (called here LR) and measure dissimilarity between images. However, another strategy proposed in the literature on re-identification consists of directly learning the distance between images using distance learning approach~\cite{bai2018group,wang2016semantic,wang2017multi}. These approaches rely on optimizing intra-/inter-class distances during model learning. Thus, as the impact of the distance definition has been shown in transfer learning LR-based approaches in this work, it could be relevant to evaluate if such impact also exists in distance learning-based approaches.

\section{Conclusion}

\label{sec:conclusion}

Recent studies on vehicle re-identification focused on the extraction of latent representation (LR) of vehicles, i.e. vectors of features extracted from the latent space of convolutional neural networks (CNN), to discriminate between vehicles on their visual appearance in order to retrieve a given vehicle.
These previous works performed the re-identification process by comparing LR of vehicles using metrics based on the Euclidean distance (or a variant), which is known to be poorly suited with high-dimensional spaces (such as CNN latent spaces). In addition, they used I2IP or I2TP for the re-identification process, using one image of a query vehicle to retrieve an image or a track (a set of images) of the probed vehicle.

In this paper, we firstly studied the impact of the metric used for the vehicle re-identification, comparing performances obtained with different metrics; we studied visual-information only re-identification processes (no extra or contextual information was used).
We tested metrics based on the minimal Euclidean distance ($MED$), the minimal cosine distance ($MCD$), and the residual of the sparse coding reconstruction ($RSCR$). We applied these metrics using features extracted from five different CNN architectures (namely ResNet18, AlexNet, VGG16, InceptionV3 and DenseNet201). 
We used the VeRi dataset to fine-tune these CNNs and to evaluate the results in I2TP. Results show a major impact of the metric on the re-identification performance. In overall, independently of the CNN used, $MCD$ metric outperforms $MED$ (mAP: [+2.02\% -- +5.79\%]). This result is of great importance since the literature mainly uses Euclidean-based distance (or a variant) during the re-identification process. Keeping the CNN providing the best performances (DenseNet201), we further evaluated the impact of the metric in other I2IP-based vehicle retrieval tasks using three other datasets (VehicleID, CompCars and BoxCars116k). In these experiments, $MCD$ also outperformed $MED$, suggesting that performance gain provided by $MCD$ could be generalized to other LR-based retrieval tasks.

In a second part, we investigated to extend the state-of-the-art I2TP to a track-to-track process (T2TP). Indeed, in real applications, users mainly operate video segments (vehicle tracks) rather than vehicle images. T2TP grounds the re-identification on the visual data available (vehicle track) and enhances the process without using additional metadata (contextual features, spatiotemporal information, etc.). We extended the metrics to measure the distance between tracks, allowing for the evaluation of T2TP and comparison with I2TP. 

Results show that T2TP outperforms I2TP for $MCD$ (mAP: +4.07\%$\pm$0.85) and for $RSCR$ (mAP: +3.37\%$\pm$3.11). T2TP combining DenseNet201 and $MCD$-based metrics shows the best performances, outperforming some of the state-of-the-art methods without integrating any additional metadata. 

To conclude, our experiments highlight the importance of the metric choice in the vehicle re-identification process. In addition, T2TP improves the vehicle re-identification performances (compared to I2TP), especially when coupled with $MCD$-based metrics. 

As practice of vehicle re-identification tends to favour queries based on tracks rather than images, we argue for considering T2TP (in addition or in replacement of I2TP) in future vehicle re-identification works.

\bibliographystyle{unsrt}  

\bibliography{biblio.bib}

\end{document}